\documentclass[letterpaper, 10 pt, conference]{ieeeconf}

\usepackage{FG2020}
 \FGfinalcopy % *** Uncomment this line for the final submission
\usepackage{times} 
\usepackage{epsfig}
\usepackage{graphicx}
\usepackage{amsmath}
\usepackage{amssymb}
\usepackage[ruled]{algorithm2e}

\usepackage{subcaption}
\usepackage{booktabs}
\usepackage{float}
\usepackage{gensymb}
\usepackage{multirow}
 \usepackage{color}

\graphicspath{{figures/}}
\newcommand{\etal}{\textit{et al}.\@ }
\newcommand{\ie}{\textit{i.e}.\@ }
\newcommand{\eg}{\textit{e.g}.\@ }

\usepackage[pagebackref=true,breaklinks=true,letterpaper=true,colorlinks,bookmarks=false]{hyperref}

\def\FGPaperID{****} % *** Enter the FG Paper ID here

\title{\LARGE \bf
Robustness Analysis of Face Obscuration
}

\author{\parbox{16cm}{\centering
    {\large Hanxiang Hao$^1$, David G\"{u}era$^1$, J\'{a}nos Horv\'{a}th$^1$, Amy R. Reibman$^2$, Edward J. Delp$^1$}\\
    {\normalsize
    $^1$ Video and Image Processing Lab (VIPER), Purdue University, West Lafayette, Indiana USA\\
    $^2$ School of Electrical and Computer Engineering, Purdue University, West Lafayette, Indiana USA}}
}

\begin{document}
\ifFGfinal
\thispagestyle{empty}
\pagestyle{empty}
\else
\author{Anonymous FG2020 submission\\ Paper ID \FGPaperID \\}
\pagestyle{plain}
\fi

\twocolumn[{
\renewcommand\twocolumn[1][]{#1}
\maketitle
\begin{center}
    \centering
	\includegraphics[width=1\linewidth]{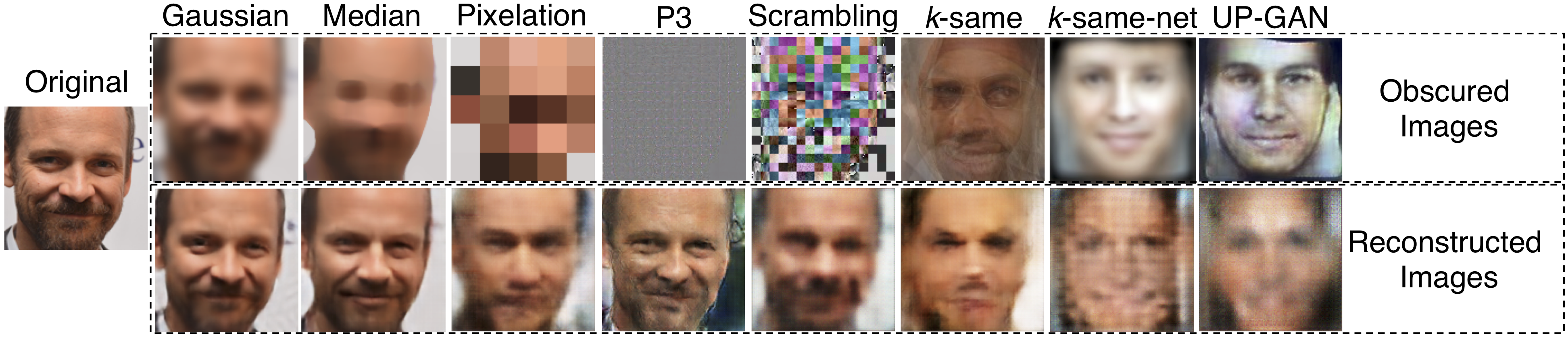}
	\captionof{figure}{Reconstruction of obscured images using Pix2Pix \cite{Isola_2016} as described in Section \ref{Reconstruction method}. Although the obscured images are hard to recognize, deep learning models can still recover the person's identity. For Gaussian, median, and P3, we can clearly recognize the person from their recovered images. }
	\label{fig:intro}
\end{center}
}]

%%%%%%%%% ABSTRACT
\begin{abstract}
Face obscuration is needed by law enforcement and mass media outlets to guarantee privacy.
Sharing sensitive content where obscuration or redaction techniques have failed to completely remove all identifiable traces can lead to many legal and social issues.
Hence, we need to be able to systematically measure the face obscuration performance of a given technique.
In this paper we propose to measure the effectiveness of eight obscuration techniques. %as shown in Figure \ref{fig:intro}.
We do so by attacking the redacted faces in three scenarios: obscured face identification, verification, and reconstruction.
Threat modeling is also considered to provide a vulnerability analysis for each studied obscuration technique. 
Based on our evaluation, we show that the $k$-same based methods are the most effective.
\end{abstract}

%%%%%%%%% BODY TEXT
\section{Introduction} \label{Introduction}
From TV news to Google StreetView, object obscuration has been used in many applications to provide privacy protection.
Law enforcement agencies use obscuration techniques to avoid exposing the identities of bystanders or officers. 
To remove identifiable information, Gaussian blurring or pixelation methods are commonly used. 
Median filtering is also used due to its simple implementation and its non-linearity, which translates into higher information distortion when compared to linear filters such as the Gaussian filter. 
These simple obscuration techniques are able to successfully prevent humans from recognizing the obscured objects.
Previous work \cite{Dufaux_2010, McPherson_2016, Sah_2017} shows that machine learning approaches can still identify these objects using the subtle information left in the obscured images. 
More robust and effective techniques have been described including $k$-same methods \cite{Newton_2005, Gross_2005, Du_2014, Meden_2018, Hao_2019} which are able to provide a secured obscuration while preserving non-identifiable information.
Reversible obscuration \cite{Ra_2013, Pares-Pulido_2014, Yuan_2015} is another type of method to prevent the leakage of privacy information from unauthorized viewers when sharing an image on social media.
This type of methods is designed to achieve privacy-preserving image sharing by encrypting the images published on the Internet.
Only the viewer with the correct decoding key is able to access the image.
% \textbf{{\color{red} The previous sentence makes no sense to me…fix it!!! Should you have a reference in this sentence?}}
In this paper, we focus on the robustness analysis of several obscuration techniques for face redaction. 
We study these obscuration methods to answer the following question: ``Is there any remaining identifiable information from the obscured faces to enable re-identification?''. 

Although several of these approaches are widely used by news outlets, social media platforms, and government agencies, their performance has not been objectively measured.
The lack of a formal study of these obscuration techniques makes it hard to evaluate the quality of redaction systems.
As shown by McPherson \etal \cite{McPherson_2016}, a deep learning model with a simple structure is able to identify individuals from their highly pixelated and blurred faces. 
This indicates that human perception is no longer the gold standard to examine the effectiveness of obscuration methods. 
To provide a better way to examine a given obscuration method, we need to consider it in a controlled environment that can determine how well identifiable information can be extracted from the obscured face.
We design three scenarios: obscured face identification, verification, and reconstruction.
Figure \ref{fig:intro} shows the results from the reconstruction attack for the eight studied obscuration methods. 
To analyze the vulnerability of these obscuration methods, we also examine multiple threat models based on an attacker's knowledge of the obscuration method used.
Our simplest threat model assumes that the attacker has no information of these obscuration methods.
In the most challenging threat scenario, we consider that the attacker knows the exact type of the obscuration method and its hyperparameters. 
These previously unexplored threat models are necessary to offer a complete vulnerability analysis under realistic situations.

The main contributions of this paper are summarized as follows. 
First, we design three attack scenarios: obscured face identification, verification, and reconstruction.
We also analyze these attacks based on two widely used deep learning models, VGG19\cite{Simonyan_2015} and ResNet50\cite{He_2016} in different threat model conditions.
Finally, we provide a comprehensive robustness analysis of eight obscuration methods. 
These methods include three traditional methods (Gaussian blurring, median blurring, and pixelation), three $k$-same based methods ($k$-same \cite{Gross_2005}, $k$-same-net \cite{Meden_2018}, and UP-GAN \cite{Hao_2019}) and two privacy-preserving image sharing methods (P3 \cite{Ra_2013} and scrambling \cite{Yuan_2015}). 

%------------------------------------------------------------------------
\section{Related Work} \label{sec:Related Work}
\textbf{Face Obscuration Methods.} 
As previously mentioned, Gaussian blurring and pixelation are frequently used in many applications.
However, these techniques are not reliable. 
As we will show in Section~\ref{sec:Experiment}, Gaussian blurring even with a large kernel size is still not able to defend against the some of our attacks.
An extreme example of blurring  to prevent information leaking is to simply gray out the entire facial region by setting all pixels in the facial area to a fixed value. 
This approach is rarely used because its visual effect is unpleasant, especially if there are many faces in the scene that need to be redacted.

To address some of these issues, $k$-same  methods \cite{Newton_2005, Gross_2005, Du_2014, Meden_2018, Hao_2019} have been proposed to balance the removal of identifiable information while preserving non-identifiable facial features.
These methods attempt to group faces into clusters based on personal attributes such as age, gender, or facial expression. 
Then, a template face for each cluster is generated.
These methods can fulfill the requirement of $k$-anonymity \cite{Samarati_2002}. 
More specifically, they are able to guarantee that any face recognition system cannot do better than $1/k$ in recognizing to whom a particular image corresponds, where $k$ is the minimum number of faces among all clusters~\cite{Gross_2005}. 
In Newton \etal \cite{Newton_2005} and Gross \etal \cite{Gross_2005}, they simply compute the average face for each cluster. 
Therefore, the obscured faces are blurry and cannot handle various facial poses. 
Du \etal \cite{Du_2014} use the active appearance model \cite{Cootes_2001} to learn the shape and appearance of faces. 
Then, they generate a template face for each cluster to produce obscured faces with better visual quality. 
A generative neural network, k-same-net, that directly generates faces based on the cluster attributes is described in \cite{Meden_2018}. 
To produce more realistic faces, generative adversarial network (GAN) \cite{Goodfellow_2014} have been used, since its discriminator is designed to guide the generator by distinguishing real faces from generated faces.
Hao \etal \cite{Hao_2019} propose a method based on conditional GAN \cite{Mirza_2014} that can generate a synthetic face given the facial landmarks and cluster attributes without the original image. 

Besides the methods above that permanently remove the identifiable information, reversible obscuration methods \cite{Ra_2013, Pares-Pulido_2014, Yuan_2015} are also needed for the purposes of privacy-preserving image sharing.
These reversible obscuration methods split the image information into two parts: 1) the public part which contains most volume, but not meaningful content and 2) a secret part that stores the image decoding key. 
Therefore, when publishing an image to social media, the public and secret parts can be stored separately to avoid the leakage of images to unauthorized viewers. 
Ra \etal \cite{Ra_2013} propose a method, P3, which is based on the JPEG encoding framework. 
They separate the DCT coefficients in the JPEG encoding process based on a predefined threshold value to generate the public and secret images. 
Yuan \etal \cite{Yuan_2015} propose a scrambling method that further reduces the data storage in the secret part. 
Instead of thresholding, they randomly flip the the sign of DCT coefficients and store the result as the public image.
For the secret part, they only need to store the random seed to recover the original image. 

\textbf{Privacy Analysis of Obscuration Methods.} 
% As mentioned in Section \ref{Introduction}, although Gaussian blurring and pixelation are widely used, these methods might still leak sensitive information. 
Dufaux and Ebrahimi, and Sah \etal~\cite{Dufaux_2010, Sah_2017} provide an analysis of the obscuration performance of simple identifiers and show the ineffectiveness of current obscuration methods. 
By using a simple deep learning model, McPherson \etal \cite{McPherson_2016} also show that obscured images still contain enough information to perform accurate identification.
They uncover the identity obscured with blurring, pixelation, and P3 methods. 
%For the $16\times16$ pixelation method, they achieve a top-5 identification accuracy of 98.75\% for the AT\&T dataset \cite{Samaria_1994} and 72.23\% for the FaceScrub dataset \cite{Ng_2014}. 
Oh \etal \cite{Oh_2016} also propose a semi-supervised model that is able to identify the face under large variations in pose. 
% Their model is based on a conditional random field (CRF), which not only infers the individual faces (unary part) but also deduces the identities based on other visible faces (pairwise part). 
% Therefore, when the unary part is weak due to the obscuration, the identifiable information from other visible faces is able to help improve the deduction of the obscured face through the connections from the CRF. 

To extend the previous literature~\cite{Dufaux_2010, Sah_2017, McPherson_2016, Oh_2016}, we first consider the face identification scenario. 
By mapping faces to known identities in different threat models, we analyze the vulnerability of each obscuration method using advanced deep learning identification methods. 
However, the requirement of known identities weakens this type of analysis, since query faces usually come from unknown identities. 
To overcome this, we provide a threat analysis under a more realistic setup: the face verification scenario.
Specifically, we want to measure the similarity of an unknown redacted face to clear target faces.
Since it allows recognizing unseen identities, this scenario is more realistic. 
Lastly, a reconstruction scenario is proposed to visualize how well we can recover the true identity using the remaining information from the obscured images.

%------------------------------------------------------------------------
\section{Proposed Method}
To evaluate the performance of the obscuration methods, we first introduce the three threat models based on the amount of knowledge about the obscuration method that is available to the attackers. 
Then we describe the three attacks: obscured face identification, verification, and reconstruction.

\subsection{Threat Modeling}
In our model, the attacker aims to identify the redacted faces based on the information still present in the obscured images.
% To be clear, we define attacker as a face recognition system that tries to reveal the identity of the obscured faces.
We design three threat models, which vary on how much information about the used obscuration approach is available to the attacker. 
% We jointly train the clear images with obscured images, since it provides us better accuracy than learning from the obscured images themselves. 
\begin{itemize}
	\item Threat model $T_1$ assumes the attacker has no information of any obscuration method, which means that the attacker is only able to learn the facial features used for identification from clear faces. 
	During the testing phase, it extracts the facial features from the obscured faces directly. 
	\item Threat model $T_2$ assumes the attacker is aware of some obscuration methods, but not the same method used in the testing phase. 
	\ie the attacker is trained on both clear and obscured images and tested with the obscured images of the obscuration methods not used in the training set. 
	This model provides more information to the attacker, since different obscuration methods may share similarities in terms of identifying facial features. 
	Introducing the obscured images serves the same role as data augmentation. 
	\item Threat model $T_3$ assumes the attacker knows the exact type of the obscuration method and its hyperparameters, like the kernel size of Gaussian blurring.
	Compared to $T_1$ and $T_2$, $T_3$ is the strongest and most realistic attack, since it provides the attacker with the most information of the obscuration method to identify identities. 
\end{itemize}

% \begin{figure}[htbp]
% 	\begin{center}
% 		\includegraphics[width=1\linewidth]{threat_model2}
% 	\end{center}
% 		\caption{Illustration of three threat models during training and testing phases.} % Not sure if need this in caption: $T_1$: train on clear images and test on obscured images. $T_2$: train on clear and obscured images and test on the obscured images from a different obscuration method using in the training set. $T_3$: train on clear and obscured images and test on the obscured images with the same obscuration method using in the training set.
% 	\label{fig:threat_model}
% \end{figure}

\subsection{Obscured Face Identification Attack} \label{Identification method}
For the obscured face identification attack, we assume a fixed number of identities.
We treat this identification problem as a classification problem where the number of classes is equal to the number of identities.
In this paper, we evaluate the performance of different obscuration methods based on different backbone deep learning models, such as VGG19 or ResNet50 in order to have a more generalizable conclusion. 

\subsection{Obscured Face Verification Attack} \label{Verification method}
The obscured face verification attack is defined as: given an obscured face and a clear face, decide if the two faces come from the same person or not. 
Previous work \cite{Dufaux_2010, Sah_2017, McPherson_2016, Oh_2016} only considers the identification scenario, which assumes all identities are in the dataset.  
However, in many cases, we cannot assume the obscured identity is in any dataset. 
For example, the attackers may want to find out if the obscured face from a TV news is a person they know. 
Therefore, face verification attack is more stringent. 

In order to solve this verification problem, we project the image into a low-dimension latent vector, where faces from the same person are closer together than faces from different people. 
Therefore, by comparing the distance of the latent vectors, we can determine if the two faces are from the same person or not. 
To improve the accuracy, we use the Additive Angular Margin loss (also known as ArcFace) \cite{Deng_2019} to obtain highly discriminative features for face recognition.
ArcFace simultaneously reduces intra-class difference and enlarge inter-class difference of the embedding vectors. 
We choose ArcFace because it yields the best facial recognition performance among the traditional softmax loss \cite{Taigman_2014, Sun_2014_b, Cao_2018}, contrastive loss \cite{Sun_2014_a, Sun_2015_a, Sun_2015_b}, triplet loss \cite{Parkhi_2015, Schroff_2015, Sankaranarayanan_2016}, and other angular space losses, like SphereFace \cite{Liu_2016, Liu_2017} and CosFace \cite{Wang_2018_a, Wang_2018_b}.
Specifically, ArcFace is designed to enforce a margin between the distance of the sample to its class center and the distances of the sample to the other centers from different classes in angular space.
% Figure \ref{fig:arcface_loss} shows the training procedure using ArcFace loss function. 
Given an input image (either clear image or obscured image), we first embed it as a low-dimension vector $\mathbf{x} \in \mathbb{R}^d$ using a deep learning model. 
Define an auxiliary projection weight $\mathbf{W} \in \mathbb{R}^{d \times n}$, where $n$ is the number of unique identities in the dataset. 
We further normalize the embedding vector and projection weight as $\hat{\mathbf{x}} = \frac{\mathbf{x}}{\Vert \mathbf{x} \Vert}$ and $\hat{\mathbf{W}} = \frac{\mathbf{W}}{\Vert \mathbf{W} \Vert}$, respectively. 
The normalized embedding vector then is projected onto $\mathbb{R}^n$ as follows
\begin{equation*}
	\hat{\mathbf{W}}^T\hat{\mathbf{x}} = \Vert \hat{\mathbf{W}} \Vert \Vert \hat{\mathbf{x}} \Vert \cos \boldsymbol{\theta} = \cos \boldsymbol{\theta}, 
\end{equation*}
where $\boldsymbol{\theta} \in \mathbb{R}^n$ is a vector of angular distance from $\hat{\mathbf{x}}$ to $\hat{\mathbf{W}}$.
The normalized embedding vector is then re-scaled by multiplying a scalar $s$ to make it distributed on a hypersphere with a radius of $s$. 
The ArcFace loss function of a single sample is then calculated using softmax cross entropy as follows 
\begin{equation*}
	L = -\log{\frac{e^{s\cos(\theta_t + m)}}{e^{s\cos(\theta_t + m)} + \sum_{j=1,j \neq t}^n e^{s\cos(\theta_j)}}}, 
\end{equation*}
where $m$ is the additive angular margin penalty between $\mathbf{x}$ and $\mathbf{W}$, $\theta_t$ is the angle of the target class of the input image.  
Note that the computation of the ArcFace loss is only used to aid the training process. 
For inference, we compute the embedding vectors from the clear face $\boldsymbol{x}_c$ and obscured face $\boldsymbol{x}_o$ using the same deep learning model. 
We then compare the angular distance after normalization to a predefined threshold value to determine the verification result. 
The threshold value can be obtained based on the value that maximizes the verification accuracy on the validation set.

% \begin{figure}[htbp]
% \begin{center}
% 	\includegraphics[width=1\linewidth]{arcface_loss2} 
% \end{center}
% 	\caption{Training procedure of the ArcFace loss function.}
% \label{fig:arcface_loss}
% \end{figure}

% \begin{figure}[htbp]
% \begin{center}
% 	\includegraphics[width=1\linewidth]{verification_inference2}
% \end{center}
% 	\caption{Inference pipeline of obscured face verification attack.}
% \label{fig:verification_inference}
% \end{figure}

\subsection{Obscured Face Reconstruction} \label{Reconstruction method}
As we will show in Section \ref{sec:Experiment}, highly obscured images still contain identifiable information. 
To examine the amount of remaining information in obscured images, we design a reconstruction attack to visualize how well we can recover the original image. 
We apply a conditional generative adversarial network, Pix2Pix \cite{Isola_2016}, to perform this image reconstruction attack. 
Given the obscured images, the generator is trained to reconstruct the clear image guided by the discriminator and the $L_2$ distance loss. 
To quantify the reconstruction performance, we compute the mean square error (MSE) over pixel-wise differences. 
We also compute the identification accuracy based on a face recognition model which is pretrained with clear images. 
This test provides us a way to quantify and visualize the amount of identifiable information leaked from the obscuration methods. 

%------------------------------------------------------------------------
\section{Experiments} \label{sec:Experiment}
In this section, we first briefly describe the obscuration methods to be evaluated. 
Then, we provide the design and result of the aforementioned attacking scenarios.

\subsection{Evaluated Methods} \label{Evaluated methods}
% \begin{figure}[htbp]
% \begin{center}
% 	\includegraphics[width=0.8\linewidth]{facescurb_obscured2}
% \end{center}
% 	\caption{Obscuration example of compared methods. First to third row: Gaussian, median and pixelation with kernel (pixel) sizes as 5, 15, 25 and 35; Fourth row: $k$-same, $k$-same-net, UP-GAN, P3 and Scrambling. }
% \label{fig:facescurb_obscured}
% \end{figure}

In this work, we propose to analyze eight obscuration methods.
These methods include three traditional methods (Gaussian blurring, median blurring, and pixelation), three $k$-same based methods ($k$-same, $k$-same-net, and UP-GAN) and two privacy-preserving image sharing methods (P3 and scrambling). 
Examples of obscured faces using these methods are shown in Figure \ref{fig:intro}. 
We use Gaussian-5 representing the experiment of Gaussian blurring with kernel size of 5. 

\textbf{Traditional obscuration methods.}
We evaluate the three obscuration methods including Gaussian blurring, median blurring and pixelation methods for four different kernel (pixel) sizes of 5, 15, 25, and 35. 
We use the OpenCV function \textit{cv2.getGaussianKernel} to compute the kernel of Gaussian blurring. 
Note that the Gaussian standard deviation is defined as
\begin{equation*}
	\sigma = 0.3 * \left(\left(\frac{w-1}{2} - 1\right) + 0.8\right), 
\end{equation*}
where $w$ is the kernel size. 
The pixelation method is implemented by image downsampling and upsampling using nearest-neighbor interpolation. 

\textbf{$k$-same based obscuration methods.}
$k$-same based methods aim to obscure identifiable information while preserving the non-identifiable information (also known as utility information). 
% Recall that $k$-same based methods are able to guarantee that no face recognition system can do better than $1/k$ in recognizing who a particular image corresponds to, where $k$ is the minimum number of faces among all clusters. 
Algorithm \ref{algo:ksame} shows the workflow of the $k$-same based methods, which is based on \cite{Gross_2005}. 
In this work, we choose $k=10$. 
We evaluate three $k$-same based methods: the original $k$-same method \cite{Gross_2005}, $k$-same-net \cite{Meden_2018}, and UP-GAN \cite{Hao_2019}. 
We model the obscuration process as follows. 

Suppose we have a clear face dataset $\mathcal{M}_c$ and an obscuration function $f$ mapping the clear image $I_c$ to the obscured image $I_o$ by $I_o=f(I_c)$. 
We use this mapping function building an obscured face dataset $\mathcal{M}_o$ based on $\mathcal{M}_c$. 
Based on \cite{Gross_2005}, we also need to assume the dataset $\mathcal{M}_c$ has no two images coming from the same identity to make Algorithm \ref{algo:ksame} $k$-anonymous. 
The $k$-same based methods require the function $f$ mapping $k$ nearest neighbors from the clear images to a single obscured image. 
For example, considering the original $k$-same method, the obscured face is obtained by averaging the $k$ nearest neighbors in the image space. 
Therefore, the $x_{1,\ldots,k}$ from Algorithm \ref{algo:ksame} in this case are the clear images.  

$k$-same-net is a generative deep learning model that generates fake faces given the cluster attributes. 
UP-GAN has similar generator architecture to $k$-same-net with the same input cluster attributes. 
However, it improves the generated image quality using its discriminator and the perceptual loss constraint. 
For both $k$-same-net and UP-GAN, the $x_{1,\ldots,k}$ from Algorithm \ref{algo:ksame} are the cluster attributes. 
Therefore, the input attribute to the models is the average of the $k$ nearest neighbors in the attribute space.  

As proposed by \cite{Hao_2019}, we choose UTKFace dataset \cite{Zhang_2017}, which contains the required utility values (age, gender, and skin tone) and facial landmarks to train $k$-same-net and UP-GAN. 
The utility values are defined as facial features that do not reveal identity, such as age, gender, skin tone, pose, and expression \cite{Hao_2019}.
% \textbf{{\color{red} Where do you CLEARLY define ``utility value''??}}
For the purpose of obscuration evaluation, we test these two methods on the FaceScrub dataset \cite{Ng_2014}, with a fixed utility values (26 years old, male, and white) and 7-point facial landmarks obtained by \textit{Dlib} toolkit \cite{King_2009}\footnote{Since we fix the utility values, in this case, $x_{1,\ldots,k}$ are the facial landmark vectors.}.  
These points include the centers of the eyes, the center of the nose, and four points around the mouth.
Note that since the FaceScrub dataset contains different faces from the same identity, the $k$-anonymity property in this case may not hold. 
\begin{algorithm}[tb]
	\SetAlgoLined
	\KwIn{Clear face dataset $\mathcal{M}_c$, privacy constant $k$ with $|\mathcal{M}_c| \geq k$}
	\KwOut{Obscured face dataset $\mathcal{M}_o$}
	$\mathcal{M}_o \leftarrow \emptyset$\;
	\For{$i \in \mathcal{M}_c$}{
	   \If{$|\mathcal{M}_c| < k$}{
	   $k = |\mathcal{M}_c|$\;
	   }
	   Select the $k$ nearest neighbors ${x_1, \ldots, x_k} \in \mathcal{M}_c$\;
	   $x_o \leftarrow \frac{\sum_{m=1}^k x_m}{k}$\;
	   Add $k$ copies of $x_o$ to $\mathcal{M}_o$\;
	   Remove $x_1, \ldots, x_k$ from $\mathcal{M}_c$\; 
	 }
	\caption{Workflow of the $k$-same based methods.}
	\label{algo:ksame}
\end{algorithm}

\textbf{Privacy-preserving image sharing methods.}
Privacy-preserving image sharing methods are designed to encrypt the content of the original image when publishing to social media. 
To recover the original images, the encrypted images need a key to decrypt the content. 
We evaluate two methods: P3 \cite{Ra_2013} and scrambling \cite{Yuan_2015}.
Both of them are based on the manipulation of DCT coefficients in the JPEG framework. 
After obtaining the DCT coefficients from $8\times8$ image patches, P3 separates the AC coefficients given a predefined threshold value. 
It then stores the coefficients that are smaller than the threshold value as the public image. 
The secret image contains the DC coefficients and the AC coefficients that are higher than the threshold value. 
In this paper, we choose the threshold value as 10. 
For the scrambling method, it first evenly and randomly flips the DCT coefficients and stores the result as the public image. 
For the secret part, it only stores the random seed. 
Therefore, it can restore the image by undoing the flipping process based on the random seed.  
In this paper, we scramble both DC and AC DCT coefficients for all YUV components, which is the high-level scrambling as proposed by \cite{Yuan_2015}.  

\subsection{Datasets} \label{dataset}
We use the FaceScrub dataset \cite{Ng_2014} which contains 106,863 face images from 530 identities. 
Therefore, the classification accuracy of randomly guessing is about 0.002. 
For the identification and verification attacks, we split the images from each identity into training, validation, and testing sets with the ratio of $6:2:2$.
For the reconstruction attack, we split the identities into three groups for the purpose of training, validation, and testing with the same ratio. 
We do so to verify if the reconstruction model is able to recover unknown identity instead of just memorizing faces. 

\subsection{Obscured Face Identification Attack} \label{Identification experiment}
\begin{table*}[htbp]
	\begin{center}
		\begin{tabular}{@{}cccccccc@{}}
			\toprule
			Method & Setting & \multicolumn{2}{c}{Threat Model $T_1$} & \multicolumn{2}{c}{Threat Model $T_2$} & \multicolumn{2}{c}{Threat Model $T_3$} \\
										  &         & VGG19 & ResNet50 & VGG19 & ResNet50 & VGG19 & ResNet50 \\ \hline
			Clear                         & -       & 0.838  & 0.890     & 0.886  & 0.884     & 0.886  & 0.884     \\ \hline
			\multirow{4}{*}{Gaussian}     & 5       & 0.787  & 0.853     & 0.829  & 0.909     & 0.891  & 0.867     \\
										  & 15      & 0.106  & 0.219     & 0.548  & 0.773     & 0.863  & 0.847     \\
										  & 25      & 0.010  & 0.030     & 0.236  & 0.573     & 0.830  & 0.819     \\
										  & 35      & 0.007  & 0.009     & 0.152  & 0.430     & 0.811  & 0.798     \\ \hline
			\multirow{4}{*}{Median}       & 5       & 0.786  & 0.855     & 0.883  & 0.907     & 0.913  & 0.907     \\
										  & 15      & 0.185  & 0.229     & 0.735  & 0.823     & 0.889  & 0.885     \\
										  & 25      & 0.025  & 0.035     & 0.357  & 0.489     & 0.856  & 0.842     \\
										  & 35      & 0.011  & 0.014     & 0.213  & 0.270     & 0.805  & 0.798     \\ \hline
			\multirow{4}{*}{Pixelation}   & 5       & 0.055  & 0.208     & 0.408  & 0.606     & 0.877  & 0.884     \\
										  & 15      & 0.004  & 0.003     & 0.008  & 0.008     & 0.651  & 0.643     \\
										  & 25      & 0.003  & 0.002     & 0.005  & 0.004     & 0.461  & 0.408     \\
										  & 35      & 0.004  & 0.002     & 0.004  & 0.005     & 0.373  & 0.323     \\ \hline
			$k$-same \cite{Gross_2005}    & 10      & 0.012  & 0.012     & 0.013  & 0.012     & 0.050  & 0.063     \\ \hline
			$k$-same-net \cite{Meden_2018}& -       & 0.091  & 0.081     & 0.091  & 0.088     & 0.095  & 0.092     \\ \hline
			UP-GAN \cite{Hao_2019}        & -       & 0.091  & 0.082     & 0.090  & 0.088     & 0.093  & 0.088     \\ \hline
			P3 \cite{Ra_2013}             & 10      & 0.001  & 0.002     & 0.002  & 0.002     & 0.678  & 0.579     \\ \hline
			Scrambling \cite{Yuan_2015}   & -       & 0.002  & 0.002     & 0.004  & 0.003     & 0.784  & 0.750     \\
			\bottomrule
		\end{tabular}
		\caption{Top-1 accuracy of the identification attack. The method \textit{Clear} means the identification of the clear image. The lower the accuracy, the better the obscuration method.}
		\label{table:Identification}	
	\end{center}
\end{table*}

\textbf{Experimental Design.} 
This attack is designed to quantify the obscuration performance in the face identification scenario. 
To have a more generalizable conclusion, we run the experiments based on two widely used backbone models, VGG19 and ResNet50. 
The input images are resized to $128 \times 128$ and the output is the softmax score for classification. 

Based on the three threat models, we design the experiments as follows. 
In the first experiment for $T_1$, the identifier is trained with the set of clear images and tested with obscured images.
In the second experiment for $T_2$, the identifier is trained on both clear and obscured images and tested with the obscured images of the obscuration method not used in the training set. 
The intuition of threat model $T_2$ is to verify if we can enforce the attacker to learn more robust features from this complex dataset. 
This can be seen as data augmentation. 
Specifically for the three traditional methods, we use the obscured images from two methods during training and use the other one for testing. 
For the $k$-same based methods and privacy-preserving image sharing methods, we train on all three traditional methods. 
Jointly training on clear and obscured images provides a better accuracy compared to learning from the obscured images themselves. 
In the third experiment for $T_3$, each identifier is trained on both clear and obscured images and tested with the obscured images using the same obscuration method. 

\textbf{Result.}
Table \ref{table:Identification} shows the identification accuracy from different obscuration methods and threat models. 
The lower the identification accuracy, the better the performance of the obscuration method. 
The results of the clear images under $T_2$ and $T_3$ are obtained by training on all three traditional methods and testing on the clear images.

We first compare the same method and same backbone model with different threat models. 
As the attackers get more information (\ie from $T_1$ to $T_3$), the identification accuracy increases. 
This means that the identifiable information left in the obscured images can still be learned by the attackers given proper training data.
For example, the accuracy of Gaussian-35 with VGG19 increases from 0.007 to 0.811 for $T_1$ and $T_3$, respectively. 
Therefore, Gaussian blurring completely fails to provide privacy for $T_3$, although visually speaking a human is not able to identify someone from the obscured images. 
A similar conclusion can be drawn for median blurring. 
Although pixelation with a large pixel size can achieve a relatively good performance, comparing the results from $T_1$ to $T_3$, the attacking accuracy still improves a lot. 
\eg for pixelation-35 with VGG19, the accuracy increases from 0.004 to 0.373, for $T_1$ and $T_3$, respectively. 
The three $k$-same based methods achieve a good obscuration performance even for $T_3$. 
For the privacy-preserving image sharing methods, although they achieve the best performance under $T_1$ and $T_2$, they still fail to provide a good obscuration under $T_3$. 
Surprisingly, even for the scrambling method which involves a random flipping process, the attackers can still extract useful features for accurate identification. 
Note that these conclusions do not change for different backbone models.

Considering $T_1$ itself, besides Gaussian-5 and median-5, all methods achieve an effective obscuration on both VGG19 and ResNet50 models. 
This means that the attackers fail to extract identifiable information from the obscured images if they solely learn from the clear image. 
% For Gaussian-5 and median-5, their blurring effect is too subtle to provide enough obscuration. 
For the three traditional methods, the obscuration performance gets better (\ie identification accuracy gets lower) as the kernel size increases. 
The original $k$-same method achieves the best obscuration performance among the three $k$-same based methods. 
For $k$-same-net and UP-GAN, since they allow the input of utility information to generate obscured faces, their obscuration performance is a little bit worse than the original $k$-same method. 
Both of the privacy-preserving image sharing methods achieve the performance of randomly guessing, which means the attackers cannot extract any identifiable information from the obscured images. 

For $T_2$, by introducing more informative training set, all traditional methods have worse performance, besides pixelation-25 and pixelation-35, which are relatively close to the results obtained from $T_1$. 
The obscuration performance of Gaussian and median blurring drops significantly (\ie the identification accuracy greatly increases). 
Because the two methods share similar blurring effects, the attackers can learn more robust features from the augmented training set. 
For the $k$-same based methods and privacy-preserving image sharing methods, compared to $T_1$, the augmented training set still does not provide useful knowledge for the attackers. 

For $T_3$, both attackers achieve the strongest attack for all cases.
Even for pixelation-35, which only contains 9 distinct pixel values, both attackers can still achieve a identification accuracy over 0.5, which is much bigger than the accuracy of randomly guessing (0.002). 
The three $k$-same based methods achieve the best obscuration performance by a great margin when compared to other methods. 
Surprisingly, the two privacy-preserving image sharing methods have a much worse performance compared to their performance in $T_1$ and $T_2$. 
Even successfully concealing the identifiable information in terms of human perception, both methods fail to provide effective obscuration. 

Therefore, based on the results from the identification attack, the $k$-same based methods achieve the best obscuration performance. 

\begin{table*}[htbp]
	\begin{center}
		\begin{tabular}{@{}cccccccc@{}}
			\toprule
			Method & Setting & \multicolumn{2}{c}{Threat Model $T_1$} & \multicolumn{2}{c}{Threat Model $T_2$} & \multicolumn{2}{c}{Threat Model $T_3$} \\
										  &         & VGG19 & ResNet50 & VGG19 & ResNet50 & VGG19 & ResNet50 \\ \hline
			Clear                         & -       & 0.984 & 0.979 & 0.984 & 0.978 & 0.984 & 0.978 \\ \hline
			\multirow{4}{*}{Gaussian}     & 5       & 0.984 & 0.978 & 0.981 & 0.979 & 0.982 & 0.977 \\ 
										  & 15      & 0.927 & 0.917 & 0.970 & 0.961 & 0.960 & 0.972 \\ 
										  & 25      & 0.674 & 0.735 & 0.880 & 0.910 & 0.968 & 0.968 \\ 
										  & 35      & 0.561 & 0.623 & 0.799 & 0.860 & 0.963 & 0.963 \\ \hline
			\multirow{4}{*}{Median}       & 5       & 0.981 & 0.975 & 0.982 & 0.978 & 0.982 & 0.979 \\ 
										  & 15      & 0.749 & 0.753 & 0.933 & 0.944 & 0.933 & 0.970 \\ 
										  & 25      & 0.618 & 0.627 & 0.830 & 0.891 & 0.876 & 0.962 \\ 
										  & 35      & 0.535 & 0.573 & 0.557 & 0.770 & 0.804 & 0.954 \\ \hline
			\multirow{4}{*}{Pixelation}   & 5       & 0.795 & 0.833 & 0.906 & 0.899 & 0.979 & 0.973 \\ 
										  & 15      & 0.504 & 0.518 & 0.755 & 0.659 & 0.889 & 0.917 \\ 
										  & 25      & 0.504 & 0.511 & 0.521 & 0.537 & 0.817 & 0.862 \\ 
										  & 35      & 0.503 & 0.505 & 0.507 & 0.517 & 0.591 & 0.780 \\ \hline
			$k$-same \cite{Gross_2005}    & 10      & 0.528 & 0.519 & 0.528 & 0.524 & 0.648 & 0.662 \\ \hline
			$k$-same-net \cite{Meden_2018}& -       & 0.553 & 0.554 & 0.550 & 0.552 & 0.511 & 0.550 \\ \hline
			UP-GAN \cite{Hao_2019}        & -       & 0.554 & 0.554 & 0.549 & 0.549 & 0.511 & 0.520 \\ \hline
			P3 \cite{Ra_2013}             & 10      & 0.501 & 0.501 & 0.501 & 0.500 & 0.878 & 0.947 \\ \hline
			Scrambling \cite{Yuan_2015}   & -       & 0.503 & 0.515 & 0.505 & 0.515 & 0.954 & 0.966 \\ 
			\bottomrule
		\end{tabular}
		\caption{AUC ROC for the verification attack. The lower the AUC, the better the obscuration method.}
		\label{table:verification}	
	\end{center}
\end{table*}

\subsection{Obscured Face Verification Attack}
\textbf{Experimental Design.}
% Recall that the verification attack is defined as: given an obscured face and a clear face, decide if the two faces come from the same person or not. 
Similarly to the experiment setting in the identification task, we resize the input image to $128 \times 128$. 
According to \cite{Deng_2019}, we choose the dimension of the embedding vector as 512 and margin $m$ as 0.5. 
However, if we use the re-scale factor $s=64$ as suggested by the original paper, we are not able to obtain a stable result. 
Therefore, after several experiments, we empirically choose the re-scale factor as $s=11$ for VGG19 and $s=8$ for ResNet50, which provides the best performance according to the validation set. 
The batch size is chosen as 128. 
We choose the stochastic gradient descent (SGD) as optimizer with a weight decay of $5e^{-4}$. 
The learning rate starts at 0.1 and is divided by 10 at the epochs of 6, 11, and 16. 
For the training of P3 and scrambling, we reduce the starting learning rate to 0.05 due to convergence issues. 
Assuming the identification task, we implement the experiments based on the three threat models. 
For the performance metric, since the face verification problem is just a binary classification problem, we choose the area under the curve (AUC) of the receiver operating characteristic (ROC) curve to examine the performance.

% \begin{figure*}
% \begin{center}
%     \centering
% 	\includegraphics[width=1\linewidth]{intro_new8}
% 	\captionof{figure}{Reconstruction of obscured images using Pix2Pix model \cite{Isola_2016}. Although the obscured images are hard to recognize, deep learning models can still recover the person's identity. For Gaussian, median, and P3, we can clearly recognize the person from their recovered images. }
% 	\label{fig:intro}
% \end{center}
% \end{figure*}

During testing we need to obtain pairs of faces with the same identity and pairs of faces with different identities. 
Due to the large number of combinations of valid pairs from the testing set, in our implementation, we only compute all valid pairs within each mini-batch (128 images which are coming from 64 identities). 
Furthermore, we run testing 10 times with different combinations of image pairs. 
The average AUC is been reported in Table \ref{table:verification}. 
The standard deviation for the tests ranges from $[0.003, 0.037]$. 
Therefore, we can directly use the average AUC to compare different experiments because of the small variation. 

\textbf{Result.}
Table \ref{table:verification} shows the verification AUC from different obscuration methods, threat models and backbone models. 
The lower the AUC, the better the performance of the obscuration method. 
We first compare the same method and same backbone model to different threat models. 
As the attackers get more information (from $T_1$ to $T_3$), the verification AUC increases. 
Take Gaussian-35 with VGG19 as an instance again. 
The AUC increases from 0.561 to 0.963 for $T_1$ and $T_3$, respectively. 
Note that the AUC for randomly guessing is 0.5. 
This means that although Gaussian-35 can successfully defend from the attack under $T_1$, after introducing the obscured data in the training set, the attackers can still extract enough identifiable information to achieve a high accuracy verification. 
For the $k$-same based methods, similar to the identification attack, they achieve a robust obscuration performance even for $T_3$. 
For the privacy-preserving image sharing methods, both of them succeed in $T_1$ and $T_2$, but fail to obscure the identities under $T_3$. 
Note that for different backbone models, although there is a small performance difference, choosing different models does not affect the conclusions reached above. 

Consider different methods with the same threat model and backbone model. 
For the traditional methods, a similar conclusion to the identification attack can be drawn.  
As the kernel (pixel) size increases, the AUC decreases for all cases, especially for pixelation-35 with VGG19 which achieves the best performance among the traditional methods. 
The $k$-same based methods achieve good results for all threat models and both attackers, which agrees with the conclusion from the identification attack.
Although the privacy-preserving image sharing methods can conceal identities well under $T_1$ and $T_2$, for the stronger $T_3$, both of them fail to provide effective obscuration.

Therefore, based on the results from the verification attack, the $k$-same based methods achieve the best obscuration performance. 

\subsection{Obscured Face Reconstruction Attack} \label{Reconstruction experiment}
\begin{table}[htbp]
	\begin{center}
		\begin{tabular}{@{}cccc@{}}
			\toprule
			Method                        & Setting & MSE$\uparrow$ & Accuracy$\downarrow$ \\ \hline
			Clear                         & -       & 0.000         & 0.849                \\ \hline
			\multirow{4}{*}{Gaussian}     & 5       & 0.000         & 0.824                \\ 
										  & 15      & 0.001         & 0.707                \\ 
										  & 25      & 0.002         & 0.519                \\ 
										  & 35      & 0.002         & 0.367                \\ \hline
			\multirow{4}{*}{Median}       & 5       & 0.001         & 0.774                \\ 
										  & 15      & 0.003         & 0.356                \\ 
										  & 25      & 0.004         & 0.152                \\ 
										  & 35      & 0.007         & 0.102                \\ \hline
			\multirow{4}{*}{Pixelation}   & 5       & 0.004         & 0.439                \\ 
										  & 15      & 0.014         & 0.043                \\ 
										  & 25      & 0.022         & 0.013                \\ 
										  & 35      & 0.031         & 0.006                \\ \hline
			$k$-same \cite{Gross_2005}    & 10      & 0.029         & 0.005                \\ \hline
			$k$-same-net \cite{Meden_2018}& -       & 0.064         & 0.018                \\ \hline
			UP-GAN \cite{Hao_2019}        & -       & 0.059         & 0.003                \\ \hline
			P3 \cite{Ra_2013}             & 10      & 0.013         & 0.339                \\ \hline
			Scrambling \cite{Yuan_2015}   & -       & 0.018         & 0.042                \\ 
			\bottomrule
		\end{tabular}
		\caption{MSE and identification accuracy of the reconstruction attack. The arrows next to \textit{MSE} and \textit{Accuracy} indicate that the higher the MSE and the lower the identification accuracy are, the better the obscuration method is.}
		\label{table:reconstruction}	
	\end{center}
\end{table}

\textbf{Experimental Design.} 
In the previous sections we show that most of the obscuration methods fail to remove all identifiable information. 
In this reconstruction attack, we try to use the remaining information from these obscured images to recover the clear image. 
If the remaining information has a strong correlation with the information from the clear image, we should be able to reconstruct the original face with a high accuracy. 
In this implementation, we choose Pix2Pix \cite{Isola_2016} which is a GAN model designed for image-to-image translation as our reconstruction model. 

Assume that the obscured images and clear images come from two distinct distributions. 
The reconstruction model aims to find a mapping function from the obscured image distribution to the clear image distribution. 
To quantify the reconstruction performance, we choose mean square error (MSE) as the metric to calculate pixel-wise distance between the clear image and the reconstructed image. 
The value range of the clear image and reconstructed image is $[0, 1]$. 
To evaluate similarity of the identifiable information from the reconstructed image and the clear image, we use the identification accuracy obtained from the ResNet50 model which is pretrained on the clear images.
This is the same setting as $T_1$, since the attacker is trained with clear images and tested with obscured images.  

\textbf{Result.} 
Figure \ref{fig:intro} shows the reconstruction results from Gaussian-25, median-25, pixelation-25, P3, scrambling, $k$-same, $k$-same-net, and UP-GAN. 
Visually, the three $k$-same based methods can successfully prevent reconstruction compared with other methods.
Although the privacy-preserving image sharing methods can prevent identification in terms of human perception, the reconstruction model can still recover the images fairly accurately, especially for P3. 
For the three traditional methods, pixelation-25 achieves a better obscuration performance compared to Gaussian-25 and median-25. 

Table \ref{table:reconstruction} shows the results of the face reconstruction attack. 
Note that setting \textit{Clear} means we input clear images to Pix2Pix model to achieve an identity mapping. 
The exact MSE for the clear image is 0.000144 and the exact MSE for Gaussian-5 is 0.000289. 
For the three traditional methods, with the kernel (pixel) size increases, the reconstruction MSE increases and the identification accuracy decreases. 
Compared to the identification attack of $T_1$, this reconstruction process can help the attackers achieve a stronger attack, since the accuracy from the reconstructed images is higher than the obscured images for most cases. 
The $k$-same based methods achieve both high MSE and low identification accuracy.
Compared to the three $k$-same methods, the two privacy-preserving image sharing methods are vulnerable to the reconstruction attack, because of their low MSE. 
% The reconstructed P3 image recovers most of the facial details, but not the background which has a solid white color. 
% This is caused by the removal of DC component from the P3 method, which represents the solid-color background.  
Therefore, as with the conclusion in the identification and verification attack, these two methods also fail to conceal identity on this reconstruction attack. 

%------------------------------------------------------------------------
\section{Conclusion}
In this paper, we propose a set of experiments to analyze the robustness of face obscuration methods. 
We provide a comprehensive analysis of eight obscuration methods: Gaussian blurring, median blurring, pixelation, $k$-same, $k$-same-net, UP-GAN, P3, and scrambling.
We examine the robustness of these methods under different attacking scenarios including identification, verification, and reconstruction with two widely used deep learning models, VGG19 and ResNet50. 
Threat modeling is also considered to evaluate the obscuration methods under different strength of attacks. 
Methods such as Gaussian blurring, median blurring, P3, and scrambling fail to provide an effective obscuration under the designed attackers, although they successfully defeat human perception. 
We also show that the $k$-same based methods can provide a secured privacy protection. 
Hence, since relying on human perception is no longer an option to guarantee privacy, the proposed set of experiments should be used to quantify and benchmark the effectiveness of any future face obscuration method. 

% \section*{Acknowledgments}
% This material is based on research sponsored by the Department of Homeland Security (DHS) under agreement number 70RSAT18FR0000161. The U.S. Government is authorized to reproduce and distribute reprints for Governmental purposes notwithstanding any copyright notation thereon. The views and conclusions contained herein are those of the authors and should not be interpreted as necessarily representing the official policies or endorsements, either expressed or implied, of DHS or the U.S. Government. 

{\small
\bibliographystyle{ieee}
\bibliography{egbib}
}

\end{document}